\documentclass[runningheads]{llncs}

\usepackage{eccv}

\usepackage{eccvabbrv}
\usepackage[noend]{algpseudocode}
\usepackage{algorithmicx}
\usepackage{caption}
\usepackage{subcaption}
\usepackage{tabularx}
\usepackage{multirow}

\usepackage{graphicx}
\usepackage{subfiles}
\usepackage{booktabs}
\usepackage{amsmath}
\usepackage{amssymb}
\usepackage{algorithm}
\usepackage{adjustbox}
\usepackage{tabularx}
\usepackage{enumitem}
\usepackage{wrapfig}

\usepackage[accsupp]{axessibility}  

\usepackage{hyperref}

\usepackage{orcidlink}

\newcommand{\mysubsection}[1]{\vspace{6pt}\noindent{\bf#1}}
\newcommand{\ours}{Parrot}
\begin{document}

\title{\ours{}: Pareto-optimal Multi-Reward Reinforcement Learning Framework for Text-to-Image Generation}

\titlerunning{Pareto-optimal Multi-Reward RL Framework for T2I Generation}

\authorrunning{Lee, Seung Hyun et al.}

\author{Seung Hyun Lee$^{1,6*}$\orcidlink{0000-0002-7773-7858}, Yinxiao Li$^{1}$\orcidlink{0009-0006-4771-3368}, Junjie Ke$^{1}$\orcidlink{0009-0000-5846-9002}, Innfarn Yoo$^{2}$\orcidlink{0000-0003-4616-4644}, \\ Han Zhang$^{3}$\orcidlink{0000-0001-7072-2189}, Jiahui Yu$^{4\text{†}}$\orcidlink{0000-0002-7085-834X}, Qifei Wang$^{2}$\orcidlink{0000-0001-7476-0190}, Fei Deng$^{2,5*}$\orcidlink{0000-0003-0690-1718}, Glenn Entis$^{1}$\orcidlink{0009-0003-0764-1353}, \\ Junfeng He$^{1}$\orcidlink{0009-0004-5465-5659}, Gang Li$^{1}$\orcidlink{0000-0002-9490-2990}, Sangpil Kim$^{6}$\orcidlink{0000-0002-7349-0018}, Irfan Essa$^{1}$\orcidlink{0000-0002-6236-2969}, Feng Yang$^{1}$\orcidlink{0000-0001-6195-2089}}

\institute{Google Research$^{1}$, Google$^{2}$, Google DeepMind$^{3}$, OpenAI$^{4}$, \\ Rutgers University$^{5}$, Korea University$^{6}$}

\maketitle
\begin{center}
    \vspace{-1.5em}
    \centering
    \captionsetup{type=figure}
    \includegraphics[width=\textwidth]{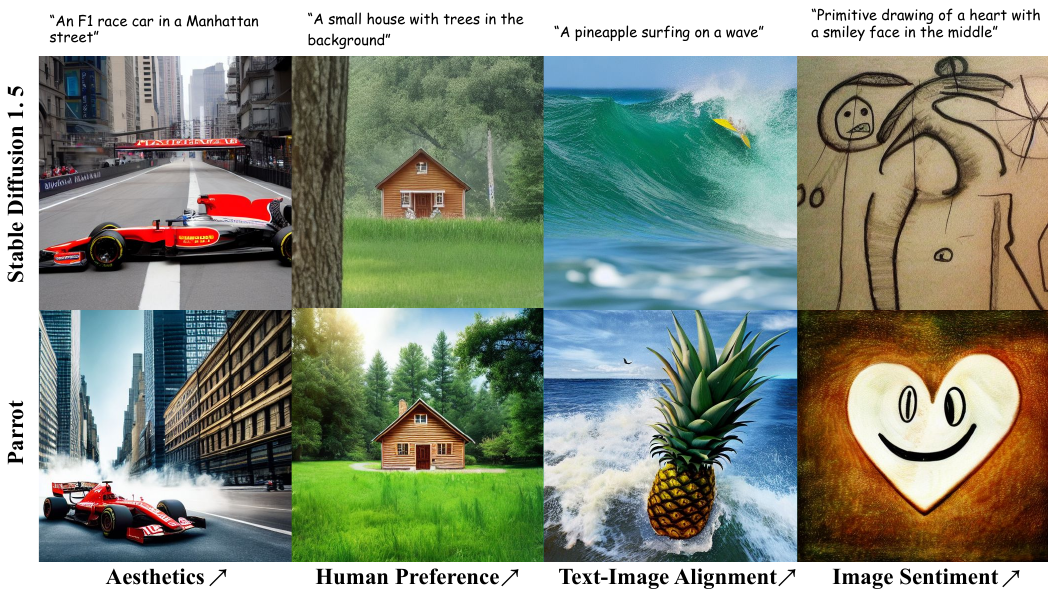}
    \vspace{-2.00em}

    \captionof{figure}{\textbf{\ours{} visual examples.} \ours{} consistently improves the quality of generated images across multiple criteria: aesthetics, human preference, text-image alignment, and image sentiment. Each column shows generated images using the same seed.}
    
    \vspace{-2.00em}
    \label{fig:fig0}
\end{center}%

\let\thefootnote\relax\footnote{\scriptsize{* This work was done during an internship at Google. \\ † This work was done during working at Google. }}

\begin{abstract}
      Recent works have demonstrated that using reinforcement learning~(RL) with multiple quality rewards can improve the quality of generated images in text-to-image (T2I) generation. However, manually adjusting reward weights poses challenges and may cause over-optimization in certain metrics. To solve this, we propose \ours{}, which addresses the issue through multi-objective optimization and introduces an effective multi-reward optimization strategy to approximate Pareto optimal. Utilizing batch-wise Pareto optimal selection, \ours{} automatically identifies the optimal trade-off among different rewards. We use the novel multi-reward optimization algorithm to jointly optimize the T2I model and a prompt expansion network, resulting in significant improvement of image quality and also allow to control the trade-off of different rewards using a reward related prompt during inference. Furthermore, we introduce original prompt-centered guidance at inference time, ensuring fidelity to user input after prompt expansion. Extensive experiments and a user study validate the superiority of \ours{} over several baselines across various quality criteria, including aesthetics, human preference, text-image alignment, and image sentiment.
\end{abstract}

\section{Introduction}
\label{sec:intro}
\noindent Despite significant advancements in text-to-image (T2I) generation ~\cite{ho2020denoising,sohl2015deep,zhou2023shifted,li2023gligen,richardson2023conceptlab}, recent work like Imagen~\cite{saharia2022photorealistic} and Stable Diffusion~\cite{rombach2022high}, still struggle to produce high quality images. Images in the first row in Fig.~\ref{fig:fig0} illustrates such quality issues in Stable Diffusion~\cite{rombach2022high}, including poor composition (\eg bad cropping), misalignment with input prompts (\eg missing objects), or overall lack of aesthetic appeal. Assessing the quality of generated images can involve various metrics such as aesthetics~\cite{ke2023vila,ke2021musiq,murray2012ava}, human preference~\cite{kirstain2023pick}, text-image alignment~\cite{radford2021learning}, and emotional appeal~\cite{serra2023emotions}.  Enhancing T2I generation across multiple quality metrics remains a challenging task.

Recent works~\cite{black2023training,fan2023dpok,lee2023aligning,hao2022optimizing} have demonstrated that incorporating quality signals as reward functions in fine-tuning T2I with reinforcement learning (RL) can improve image quality. For instance, Promptist~\cite{hao2022optimizing} fine-tunes prompt expansion model using RL with the sum of aesthetics and text-image alignment scores. However, the simple weighted sum approach may not effectively handle trade-offs among multiple quality metrics. As the number of rewards increases, manually adjusting reward weights becomes impractical. Moreover, optimizing one quality metric may inadvertently compromise others, as the model might prioritize aesthetics over relevance to the input prompt. Additionally, the trade-off for different reward is not controllable after training.

To address these challenges, we propose \textbf{\ours{}}, a \textbf{Pa}reto-optimal multi-\textbf{r}eward \textbf{r}einf\textbf{o}rcement learning algorithm to improve \textbf{t}ext-to-image generation. Unlike previous approaches that treat T2I reward optimization as a single objective optimization problem, \ours{} tackles this challenge through multi-objective optimization and introduces an effective multi-reward optimization strategy to achieve Pareto optimal approximation. Intuitively, each generated sample in a batch embodies a distinctive trade-off among various quality rewards, with some samples exhibiting superior trade-offs compared to others. Instead of updating gradients using all batch samples, \ours{} uses non-dominated points~\cite{miettinen1999nonlinear} which have better trade-offs. Consequently, \ours{} automatically learns from the optimal trade-off among different rewards. Moreover, \ours{} learns reward-specific preference prompts, which can be utilized individually or in combination to control the trade-off among different rewards during inference time. Unlike prior work, which either solely fine-tunes the T2I model~\cite{fan2023dpok} or only tunes the prompt expansion network while freezing the T2I model~\cite{hao2022optimizing}, we employ the \ours{} multi-reward optimization algorithm to jointly optimize both the T2I model and the prompt expansion network. This collaborative optimization unlocks the full potential of \ours{} by encouraging both more details from added context from the prompt expansion model, and the overall quality improvement on the T2I generation. During inference, we further introduce \textit{original prompt-centered guidance} to ensure the output image is relevant to input prompts after prompt expansion.

In summary, our contributions can be listed as follows:
\begin{itemize}[nolistsep,noitemsep]
   \item We propose \ours{}, a novel multi-reward optimization algorithm for T2I RL fine-tuning. Leveraging batch-wise Pareto-optimal selection, it effectively optimizes multiple T2I rewards, enabling collaborative improvement in aesthetics, human preference, image sentiment, and text-image alignment and also allowing to control the trade-off of different rewards using a reward related prompt during inference.
   \item We show the advantage of jointly optimizing both the prompt expansion network and the T2I model, which has never been explored before.
   \item We introduce original prompt-centered guidance during inference time after prompt expansion, ensuring better alignment with the original prompt while enriching image details.
   \item Extensive results and a user study validate that \ours{} outperforms several baseline methods across various quality criteria.
 \end{itemize}

\section{Related Work}
\label{sec:related}

\mysubsection{T2I Generation:} The goal of T2I generation is to create an image given an input text prompt. Several T2I generative models have been proposed and have demonstrated promising results~\cite{chang2023muse,yu2023magvit,ramesh2022hierarchical,saharia2022photorealistic,jeong2023power,han2023svdiff,lee2023soundini,saharia2022palette,kawar2023imagic, dai2023emu}. Stable Diffusion~\cite{rombach2022high} shows impressive generation performance in T2I generation, leveraging latent text representations from LLMs. Despite substantial progress, the images generated by those models still exhibit quality issues, such as bad cropping or misalignment with the input texts.

\mysubsection{RL for T2I Fine-tuning:} Starting by Fan~\emph{et al.}~\cite{fan2023optimizing} to explore RL fine-tuning for T2I models, following works~\cite{black2023training,fan2023dpok,dong2023raft,he2023learning, deng2024prdp} have explored RL fine-tuning technique for T2I diffusion model, showcasing superior performance for human preference learning. DPOK~\cite{fan2023dpok} improves quality through RL using ImageReward~\cite{xu2023imagereward} score as a reward with a few prompts. In addition to fine-tuning the T2I model directly using RL, Promptist~\cite{hao2022optimizing} fine-tunes the prompt expansion model by using a simple sum of aesthetic and text-image alignment scores as reward. DRaFT~\cite{clark2023directly} proposed not only differentiable rewards for efficient fine-tuning but also effectiveness of using linear summation of multi-rewards. These methods treat T2I RL as a single-objective optimization problem, while \ours{} employs multi-objective optimization. Additionally, prior approaches either fine-tune the T2I model or the prompt expansion model while freezing the other. In contrast, \ours{} proposes joint optimization of the prompt expansion model and the T2I model using multi-reward RL to foster better collaboration. 

\mysubsection{Multi-objective Optimization:} Multi-objective optimization problem involves optimizing multiple objective functions simultaneously. The scalarization technique~\cite{mannor2001steering,tesauro2007managing} formulates multi-objective problem into single-objective problems with the weighted sum of each score, which requires pre-defined weights for each objective. Rame~\textit{et al.}~\cite{rame2023rewarded} proposed weighted averaging method to find Pareto frontier, leveraging multiple fine-tuned models. Lin \etal~\cite{lin2022pareto} proposes to learn a set model to map trade-off preference vectors  to their corresponding Pareto solutions. Inspired by this, \ours{} introduces a language-based preference vector constructed from task identifiers for each reward, then encoded by the text-encoder. In the context of multi-reward RL for T2I diffusion models, Promptist~\cite{hao2022optimizing} uses a simple weighted sum of two reward scores. This approach requires manual tuning of the weights, which makes it time-consuming and hard to scale when the number of rewards increases.

\mysubsection{Generated Image Quality:} The quality assessment of images generated by T2I models involves multiple dimensions, and various metrics have been proposed. In this paper, we consider using four types of quality metrics as rewards: aesthetics, human preference, text-image alignment, and image sentiment. Aesthetics captures the overall visual appealingness of the image, and it is learned using human ratings for aesthetics in real images ~\cite{murray2012ava,ke2021musiq,ke2023vila,fang2020perceptual,hosu2020koniq,ying2020patches,tu2022maxvit}. Human preferences, rooted the concept of learning from human feedback~\cite{bai2022training,ouyang2022training,wu2023human}, involves gathering preferences at scale by having raters to compare generated images~\cite{kirstain2023pick, xu2023imagereward}. Text-image alignment measures the extent to which the generated image aligns with the input prompt, CLIP~\cite{radford2021learning} score is often employed, measuring the cosine distance of between contrastive image embedding and text embedding. Image sentiment is important for ensuring the generated image evokes positive emotions in the viewer. Serra \etal~\cite{serra2023emotions} predict average polarity of sentiments an image elicits and learn estimates for positive, neutral, and negative scores. In \ours{}, we use its positive score as a reward for positive emotions. 

\section{Preliminary}
\label{method:preliminary}
\mysubsection{Diffusion Probabilistic Models:} Diffusion probabilistic models~\cite{ho2020denoising} generate the image by gradually denoising a noisy image. Specifically, given a real image ${\bf x}_0$ from the data distribution ${\bf x}_0 \sim q({\bf x}_0)$, the forward process $q({\bf x}_t| {\bf x}_0, c)$ of diffusion probabilistic models produce a noisy image ${\bf x}_t$, which induces a distribution $p({\bf x}_0, c)$ conditioned on text prompt $c$. In classifier-free guidance~\cite{ho2022classifier}, denoising model predicts noise $\bar{\epsilon}_\theta$ with a linear combination of the unconditional score estimates $\epsilon_\theta({\bf x}_t, t)$ and the conditional score estimates $\epsilon_\theta({\bf x}_t, t, c)$ as follows:
\begin{equation}
\label{conditional_score}
    \bar{\epsilon}_\theta=w \cdot \epsilon_\theta({\bf x}_t, t, c) + (1-w)\cdot\epsilon_\theta({\bf x}_t, t, \text{null}), 
\end{equation}
where $t$ denotes diffusion time step, the $\text{null}$ indicates a null text and $w$ represents the guidance scale of classifier-free guidance where $w \geq 1$. Note that $\epsilon_\theta$ is typically parameterized by the UNet~\cite{ronneberger2015u}.

\mysubsection{RL-based T2I Diffusion Model Fine-tuning: } Given a reward signal from generated images, the goal of RL-tuning for T2I diffusion models is to optimize the policy defined as one denoising step of T2I diffusion models. In particular, Black~\textit{et al.}~\cite{black2023training} apply policy gradient algorithm, which regards the denoising process of diffusion models as a Markov decision process~(MDP) by performing multiple denoising steps iteratively. Subsequently, a black box reward model $r(\cdot, \cdot)$ predicts a single scalar value from sampled image ${\bf x}_0$. Given text condition $c \sim p(c)$ and image ${\bf x}_0$, objective function $\mathcal{J}$ can be defined to maximize the expected reward as follows:
\begin{equation}
    \mathcal{J}_\theta = \mathbb{E}_{p(c)} \mathbb{E}_{p_\theta({\bf x}_0 | c)} [r({\bf x}_0, c)],
\end{equation}
where the pre-trained diffusion model $p_\theta$ produces a sample distribution $p_\theta({\bf x}_0 | c)$ using text condition $c$. Modifying this equation, Fan~\textit{et al.}~\cite{fan2023dpok} demonstrate that the gradient of objective function $\nabla \mathcal{J}_\theta$ can be calculated through gradient ascent algorithm without using the gradient of reward model as follows:
\begin{equation}
    \begin{aligned}
        \nabla \mathcal{J}_\theta = \mathbb{E} [ r({\bf x}_0, c) \sum^T_{t=1} \nabla_\theta \log p_\theta({\bf x}_{t-1} | c, t, {\bf x}_t) ],
    \end{aligned}
\end{equation}
where $T$ denotes the total time step of the diffusion sampling. With parameters $\theta$, the expectation value can be taken over the trajectories of diffusion sampling.

\section{Method}
\label{sec:method}

\subsection{\ours{} Overview}
\label{method:llm}

\begin{figure*}[ht]
    \centering
    \includegraphics[width=\textwidth]{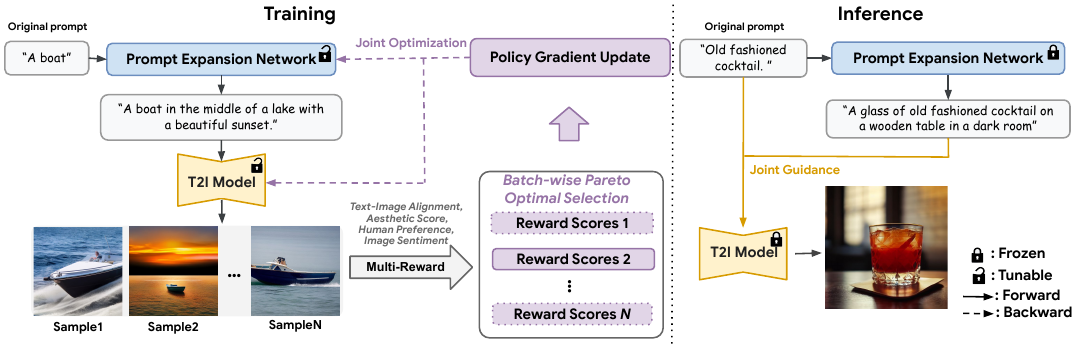}
    \vspace{-2.0em}
    \caption{Overview of \textbf{\ours{}}. During the training, $N$ images are sampled from the T2I model using the expanded prompt from the prompt expansion network. Multiple quality rewards are calculated for each image, and the Pareto-optimal set is identified using the non-dominated sorting algorithm. These optimal images are then used to perform policy gradient update of the parameters of T2I model and prompt expansion network jointly. During the inference, both the original prompt  and the expanded prompt are provided to the T2I model, enabling better faithfulness while adding detail. }
    
    \label{fig:my_label}
\end{figure*}

\noindent Fig.~\ref{fig:my_label} shows the overview of \ours{}, which consists of the prompt expansion network (PEN) $p_\phi$ and the T2I diffusion model $p_\theta$. The PEN is first initialized from a supervised fine-tuning checkpoint on demonstrations of prompt expansion pairs, and the T2I model is initialized from pretrained diffusion model. Given the original prompt $c$, the PEN generates an expanded prompt $\hat{c}$, and the T2I model  generates images based on this expanded prompt. During the multi-reward RL fine-tuning, a batch of $N$ images is sampled, and multiple quality rewards are calculated for each image, encompassing aspects like text-image alignment, aesthetics, human preference, and image sentiment. Based on these reward scores, \ours{} identifies the batch-wise Pareto-optimal set using a non-dominated sorting algorithm. This optimal set of images is then used for joint optimization of the PEN and T2I model parameters through RL policy gradient update.  During inference, \ours{} leverages both the original prompt and its expansion, striking a balance between maintaining faithfulness to the original prompt and incorporating additional details for higher quality.

\subsection{Batch-wise Pareto-optimal Selection}
\label{method:rl}
\noindent Li~\textit{et al.}~\cite{lin2022pareto} has demonstrated that using batchwise Pareto-set learning and selecting good samples in a batch can approximate Pareto-optimality across multiple objectives. Backed up by this theory, we propose to select non-dominated points in a batch for policy gradient update in RL to achieve Pareto-optimality. Algorithm~\ref{alg} outlines the procedure of \ours{}. Rather than updating the gradients using all images, \ours{} focuses on high-quality samples, considering multiple quality  rewards in each mini-batch. In the multi-reward RL, each sample generated by the T2I model presents distinct trade-offs for different rewards. Among these samples, a subset with varied optimal trade-offs across multiple objectives, also known as the Pareto set, exists. For a Pareto-optimal sample, none of its objective values can be further improved without damaging others. In other words, the Pareto-optimal set is not dominated by any data points, also known as \textit{the non-dominated set}. To achieve a Pareto-optimal solution with text-to-image generation diffusion model, \ours{} selectively uses data points from the non-dominated set using non-dominated sorting algorithm. This naturally encourages the T2I model to produce Pareto-optimal samples with respect to the multi-reward objectives.

\mysubsection{Reward-specific Preference:} Inspired by the use of preference information in multi-objective optimization~\cite{lin2022pareto}, \ours{} incorporates the preference information through reward-specific identifiers. This enables \ours{} to automatically determine the importance for each reward objective. Concretely, we enrich the  expanded prompt $\hat{c}$ by prepending reward-specific identifier ~\textit{``$<$reward k$>$''} for $k$-th reward. Based on this reward-specific prompt, $N$ images are generated and are used for  maximizing the corresponding $k$-th reward model during gradient update. At inference time, a concatenation of all the reward identifiers ~\textit{``$<$reward 1$>$,...,$<$reward~\textit{K}$>$}'' is used for image generation.

\begin{algorithm*}[t!]
\caption{Parrot: Pareto-optimal Multi-Reward RL}
\label{alg}
\caption*{\scriptsize \textbf{Algorithm} Parrot: Pareto-optimal Multi-Reward RL}
\scriptsize \textbf{Input:} Prompt $c$, Batch size $N$, Total iteration $E$, the number of rewards: $K$, Prompt expansion network $p_\phi$, T2I diffusion model: $p_\theta$, Total diffusion time step $T$, Non-dominated set: $\mathcal{P}$

\begin{minipage}{0.6\textwidth}
\label{alg}
\begin{algorithmic}
\scriptsize
\For{$e=1$ to $E$}
    \State Sample text prompt $c \sim p(c)$
    \For{$k=1$ to $K$}
        \State Expand text prompt $\hat{c} \sim p_\phi(\hat{c} |c)$
        \State Prepend reward-specific tokens \textit{``$<$reward  k$>$''} to $\hat{c}$
        \State Sample a set of images $\{{\bf x}_0^1,...,{\bf x}_0^N\} \sim p_\theta(x_0|\hat{c})$
        \State A set of reward vector $\mathcal{R}=\{R_1, ... ,R_N\}$ 
        \State $\mathcal{P} \gets$ \Call{NDSet}{$\{{\bf x}_0^1,...,{\bf x}_0^N\}$}
        \State $\nabla \mathcal{J}_\phi \mathrel{+}=  -r_k({\bf x}_0^j, \hat{c}) \times \nabla \log p_\phi(\hat{c} | c)$
    \EndFor
    \State Update the gradient $p_\theta$ from Eq.~\ref{eq:pareto}
    \State Update the gradient $p_\phi$
\EndFor
\end{algorithmic}
\end{minipage}%
\begin{minipage}{0.4\textwidth}
\begin{algorithmic}
\scriptsize
\Function{NDSet}{$\{{\bf x}_0^1,...,{\bf x}_0^N\}$}
    \State $\mathcal{P} \gets \emptyset$
    \For{$i=1$ to $N$}
        \State dominance $\gets$ True
        \For{$j=1$ to $N$}    
            \If{${\bf x}_0^j$ dominates ${\bf x}_0^i$}
                \State dominance $\gets$ False
            \EndIf
        \EndFor
        \If{dominance is True}
            \State Add $i$ to $\mathcal{P}$
        \EndIf
    \EndFor
    \State \textbf{return} $\mathcal{P}$
\EndFunction
\end{algorithmic}
\label{alg}
\end{minipage}
\\\hrulefill
\textbf{Output:} Fine-tuned diffusion model $p_\theta$, prompt expansion network $p_\phi$
\label{alg}
\end{algorithm*}

\mysubsection{Non-dominated Sorting:} \ours{} constructs Pareto set with non-dominated points based on trade-offs among multiple rewards. These non-dominated points are superior to the remaining solutions and are not dominated by each other. Formally, the dominance relationship is defined as follows: the image ${\bf x}_0^a$ dominates the image ${\bf x}_0^b$, denoted as ${\bf x}_0^b < {\bf x}_0^a$, if and only if $R_i({\bf x}_0^b) \leq R_i({\bf x}_0^a)$ for all $i \in { 1, 2,..., m }$, and there exists $j \in { 1, 2, ..., m }$ such that $R_j({\bf x}_0^b) < R_j({\bf x}_0^a)$. For example, given the $i$-th generated image ${\bf x}_0^i$ in a mini-batch, when no point in the mini-batch dominates ${\bf x}_0^i$, it is referred to as a non-dominated point.

\mysubsection{Policy Gradient Update:} We assign a reward value of zero to the data points not included in non-dominated sets and only update the gradient of these non-dominated data points as follows:  
\begin{equation}
    \begin{aligned}
        \nabla \mathcal{J}_\theta = \sum_{k=1}^K {\frac{1} {n(\mathcal{P})}} \sum_{i=1, {\bf x}_0^i \in \mathcal{P}}^N   \sum^T_{t=1} r_k({\bf x}_0^i, c_k) \times \nabla_\theta \log p_\theta({\bf x}_{t-1}^i | c_k, t, {\bf x}_t^i), 
    \end{aligned}
    \label{eq:pareto}
\end{equation}
where $i$ indicates the index of images in mini-batches, and $\mathcal{P}$ denotes batch-wise a set of non-dominated points. $K$ and $T$ are the total number of reward models and total diffusion time steps, respectively. The same text prompt is used when updating the diffusion model in each batch.

\subsection{Original Prompt Centered Guidance}
\label{method:con}
\noindent While prompt expansion enhances details and often improves generation quality, there is a concern that the added context may dilute the main content of the original input. To mitigate this during the inference, we introduce original prompt-centered guidance. When sampling conditioned on the original prompt, the diffusion model $\epsilon_\theta$ typically predicts noises by combining the unconditioned score estimate and the prompt-conditioned estimate. Instead of relying solely on the expanded prompt from PEN, we propose using a linear combination of two guidances for T2I generation: one from the user input and the other from the expanded prompt. The strength of the original prompt is controlled by guidance scales $w_1$ and $w_2$. The noise $\bar{\epsilon}_\theta$ is estimated, derived from Eq.~\ref{conditional_score}, as follows:

\begin{align}
     \bar{\epsilon}_\theta= w_1 \cdot \epsilon_\theta({\bf x}_t, t, c) + (1-w_1-w_2)\cdot\epsilon_\theta({\bf x}_t, t, \text{null}) +  w_2 \cdot \epsilon_\theta({\bf x}_t, t, \hat{c}),
     \label{eq:query}
\end{align}
where null denotes a null text.

\section{Experiments}
\label{sec:result}

\subsection{Experiment Setting}
\label{sec:setting}

\mysubsection{Dataset:} The PEN is first supervised fine-tuned on a large-scale text dataset named the Promptist~\cite{hao2022optimizing}, which has 360K constructed prompt pairs for original prompt and prompt expansion demonstration. The original instruction~{``Rephrase''} is included per pair in Promptist. We modify the instruction into~{``Input: $<$original prompt$>$. This is a text input for image generation. Expand prompt for improving image quality. Output: ''}. Subsequently, we use the RL tuning prompts~(1200K) from Promptist for RL training of the PEN and T2I model.

\begin{figure}[!tp]
    \centering
    \includegraphics[width=0.98\linewidth]{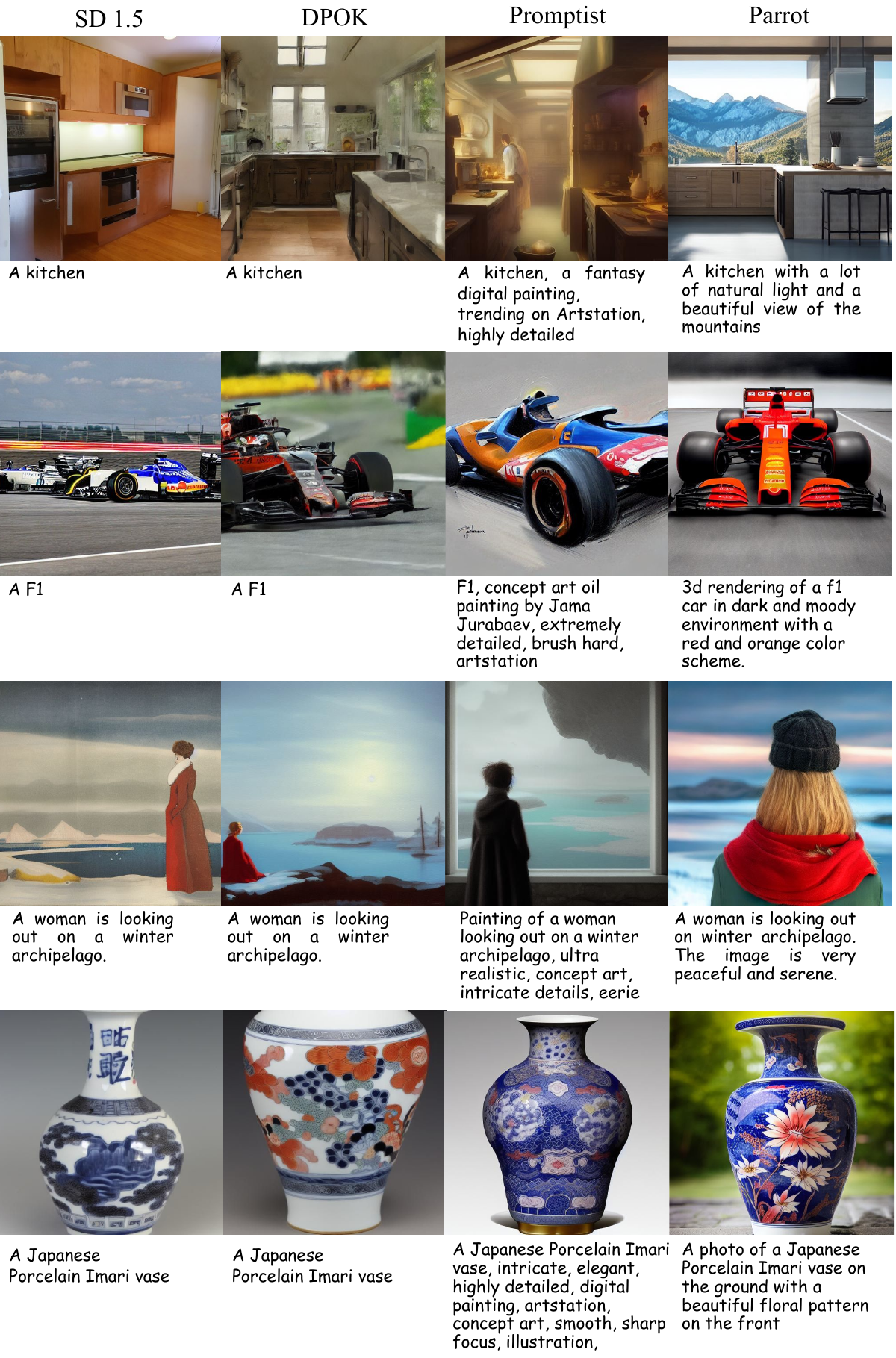}
    \caption{\textbf{Comparison of \ours{} and diffusion-based RL baselines.} From left to right, we provide results of Stable diffusion 1.5~\cite{rombach2022high}~(1st column), DPOK~\cite{fan2023dpok}~(2nd column) with the weighted sum, Promptist~\cite{hao2022optimizing}~(3rd column), and Parrot~(4th column). }
    \label{fig:joint}
\end{figure}

\mysubsection{T2I Model:} Our T2I model is based on the JAX version of Stable Diffusion 1.5~\cite{rombach2022high} pre-trained with the LAION-5B~\cite{schuhmann2022laion} dataset. We conduct experiments on a machine equipped with 16 NVIDIA RTX A100 GPUs. DDIM~\cite{song2020denoising} with 50 denoising steps is used, and the classifier-free guidance weight is set to 5.0 with the resolution 512$\times$512. Instead of updating all layers, we specifically update the cross-attention layer in the Denoising U-Net. For optimization, we employ the Adam~\cite{kingma2014adam} optimizer with a learning rate of 1$\times$$10^{-5}$.

\mysubsection{Prompt Expansion Network:} For prompt expansion, we use PaLM 2-L-IT~\cite{anil2023palm}, one of the PaLM2 variations, which is a multi-layer Transformer~\cite{vaswani2017attention} decoder with casual language modeling. We optimize LoRA~\cite{hu2021lora} weights for RL-based fine-tuning. The output token length of the PEN is set  to $77$ to match the maximum number of token length for Stable Diffusion. For original prompt-centered guidance, we set both $w_1$ and $w_2$ to 5 in Eq.~\ref{eq:query}.

\begin{figure}[!tp]
    \centering
    \includegraphics[width=\linewidth]{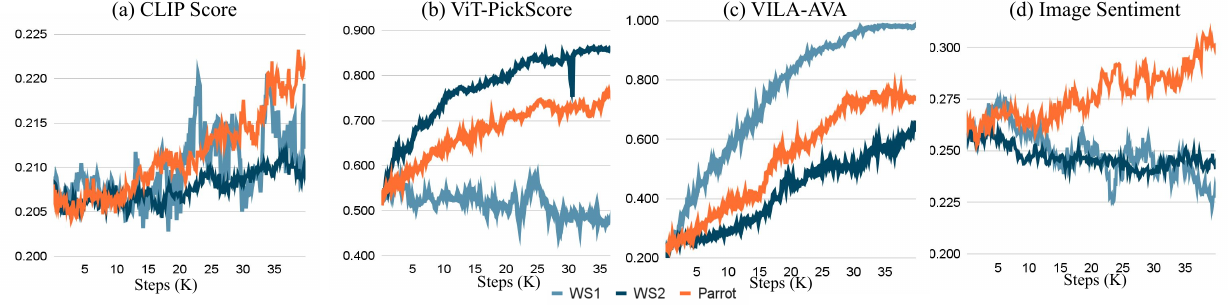}
    \vspace{-2.0em}
    \caption{\textbf{Training curve for fine-tuning on weighted sum and \ours{}.} For weighted sum, WS1 denotes $\{ 0.7, 0.1, 0.1, 0.1\}$ and WS2 denotes $\{ 0.25, 0.25, 0.25, 0.25\}$ for aesthetics, human preference, text-image alignment and image sentiment. Using weighted sum leads to decrease in human preference score and image sentiment score despite an improvement in the aesthetic score. In contrast, \ours{} exhibits stable increases across all metrics.}
    \label{fig:pareto}
\end{figure}

\mysubsection{Reward Models:} We incorporate four quality signals as rewards: Aesthetics, Human preference, Text-Image Alignment, Image Sentiment. For aesthetics, we use the VILA-R~\cite{ke2023vila} pre-trained with the AVA~\cite{murray2012ava} dataset. For human preference, we train a ViT-B/16~\cite{dosovitskiy2020image} using the Pick-a-Pic~\cite{kirstain2023pick} dataset, which contains 500K examples for human feedback in T2I generation. The ViT-B/16 image encoder consists of 12 transformer layers, and the image resolution is 224$\times$224 with a patch size of 16$\times$16.  For text-image alignment, we use CLIP~\cite{radford2021learning} with the image encoder ViT-B/32. For image sentiment, we use the pre-trained model from ~\cite{serra2023emotions}, which outputs three labels: positive, neutral, negative. We use the positive score ranging from 0 to 1 as the sentiment reward.

\subsection{Qualitative Analysis}
\label{sec:res_qual}
\mysubsection{Comparison with Baselines:} Fig.~\ref{fig:joint} shows the visual comparison of \ours{} and multiple baselines. We include results from Stable Diffusion 1.5, DPOK~\cite{fan2023dpok} with a weighted sum of rewards, Promptist~\cite{hao2022optimizing}, and \ours{}. DPOK exclusively fine-tunes the T2I model, while Promptist focuses on fine-tuning only the prompt expansion network. Parrot shows visually better images, particularly in aspects like color combination, cropping, perspective, and fine details in the image. This improvement can be attributed to \ours{}'s T2I model being fine-tuned together with the prompt expansion model that incorporates aesthetic keywords during training. \ours{} generates results that are more closely aligned with the input prompt, as well as more visually pleasing. 

\mysubsection{Weighted sum vs. Parrot:} Fig.~\ref{fig:pareto} shows the training curve comparison of \ours{} and using a linear combination of the reward scores. Each subgraph represents a reward. WS1 and WS2 denote two different weights with multiple reward scores. WS1 places greater emphasis on the aesthetic score, while WS2 adopts balanced weights across aesthetics, human preference, text-image alignment, and image sentiment. Employing the weighted sum of multiple rewards leads to a decrease in the image sentiment score, despite notable enhancements in aesthetics and human preference. In contrast, \ours{} consistently exhibits improvement across all metrics. 

\begin{table}[!tp]
    \centering
    \setlength{\tabcolsep}{4pt}
    \scriptsize
    \begin{tabularx}{\linewidth}{lcccccc}
        \toprule
        \multirow{2}{*}{Model} & \multicolumn{5}{c}{Quality Metrics} \\
        \cmidrule(lr){2-6}
         & TIA ($\uparrow$) & Aesth. ($\uparrow$) & HP ($\uparrow$) & Sent. ($\uparrow$) & Average ($\uparrow$) \\
        \midrule
        SD 1.5~\cite{rombach2022high} & 0.2322 &  0.5755 & 0.1930 & 0.3010  & 0.3254 \\
        DPOK~\cite{fan2023dpok} (WS) & 0.2337 & 0.5813 & 0.1932 & 0.3013 & 0.3273~(+0.58\%) \\
        Parrot w/o PE & 0.2355 & 0.6034 & 0.2009 & 0.3018 & 0.3354~(+3.07\%)\\
        Parrot T2I Model Tuning Only & \textbf{0.2509} & \textbf{0.7073} & \textbf{0.3337} & \textbf{0.3052} & \textbf{0.3992~(+22.6\%)} \\
        \midrule
        Promptist~\cite{hao2022optimizing} & 0.1449 & 0.6783 & 0.2759 & 0.2518 & 0.3377~(+3.77 \%)\\
        Parrot with HP Only & 0.1543 & 0.5961 & \textbf{0.3528} & 0.2562 & 0.3398~(+4.42 \%) \\
        Parrot PEN Tuning Only & 0.1659 & 0.6492 & 0.2617 & 0.3131 & 0.3474~(+6.76 \%) \\
        Parrot w/o Joint Optimization & 0.1661 & 0.6308 & 0.2566 & 0.3084 & 0.3404~(+4.60 \%) \\
        Parrot w/o ori prompt guidance & 0.1623 & 0.7156 & 0.3425 & 0.3130 & 0.3833~(+17.8 \%) \\
        Parrot & \textbf{0.1667} & \textbf{0.7396} & 0.3411 & \textbf{0.3132} & \textbf{0.3901~(+19.8 \%)} \\
        \bottomrule
    \end{tabularx}
    \caption{Quantitative comparison between Parrot and alternatives on the Parti dataset~\cite{yu2022scaling}. Abbreviations: WS - Weighted Sum; PE - Prompt Expansion; TIA - Text-Image Alignment; Aesth. - Aesthetics; HP - Human Preference; Sent. - Image Sentiment. TIA score is measured against the original prompt without expansion.}
    \vspace{-2.0em}
    \label{tab:mul}
\end{table}

\subsection{Quantitative Evaluation}
\mysubsection{Comparison with Baselines:} Table~\ref{tab:mul} presents our results of the quality score across four quality rewards: text-image alignment score, aesthetic score, human preference score, and emotion score. The first group shows methods without prompt expansion, and the second group compares methods with expansion. The prompt expansion and T2I generation are performed on the PartiPrompts~\cite{yu2022scaling}. Using a set of 1632 prompts, we generate 32 images for each text input and calculate the average for each metric.

~\textit{w/o PE} indicates the generation of images solely based on original prompt without expansion. Note that~\textit{w/o PE} is not the same as~\textit{Parrot T2I Model Tuning Only}. The former takes a model trained with PEN and removes it during inference, which results in a notable disparity between prompts used in training and testing. The latter trains with the multi-objective RL using only the T2I model, and it shows substantial improvement upon DPOK~\cite{fan2023dpok}~(weighted sum) with balanced weights of $\{0.25, 0.25, 0.25, 0.25\}$. This shows that the proposed RL method is indeed effective in multi-objective optimization.


For Promptist~\cite{hao2022optimizing}, we generate prompt expansion from their model. \textit{Parrot with HP Only} shows solely using a human prefernce reward leads to decline in others.~\textit{Parrot PEN Tuning Only} shows suboptimal aesthetics and human preference.~\textit{Parrot w/o Joint Optimization} shows suboptimal results than Parrot which demonstrates the necessity of jointly optimizing PEN and T2I models.

Our method outperforms both compared methods in aesthetics, human preference and sentiment scores. The text-image alignment score is measured with the original prompt before expansion for fair comparison. As a result, the group without prompt expansion generally shows a higher text-image alignment score. \ours{} shows better text-image alignment in each subgroup. 


\begin{table}[htp!]
    \centering
    \scriptsize 
    \begin{tabularx}{\linewidth}{lX}
        \toprule
        \textbf{Area} & \textbf{Question}  \\
        \midrule
        Aesthetics & \texttt{``Which image shows better aesthetics without blurry texture, unnatural focusing, and poor color combination?''} \\
        Human Preference & \texttt{``Which generated image do you prefer?''} \\
        Text-Image Alignment & \texttt{``Which image is well aligned with the text?''} \\
        Image Sentiment & \texttt{``Which image is closer to amusement, excitement, and contentment?''} \\
        \bottomrule
    \end{tabularx}
    \caption{Questions for user study. For performing user study, we carefully design questions suitable to each quality.}
    \label{tab:mul2}
\end{table}

\begin{figure}[t!]
    \centering
    \begin{minipage}{0.45\textwidth}
        \centering
        \includegraphics[width=\linewidth]{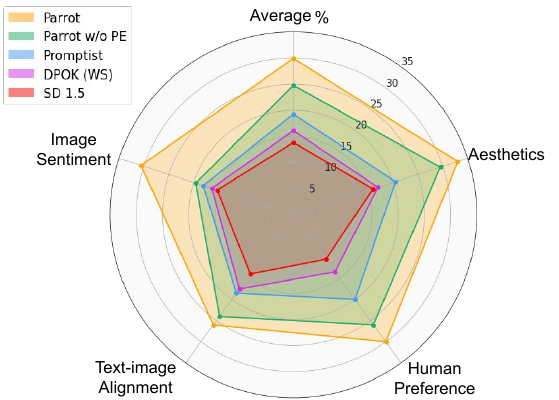}
        \caption{User study results on PartiPrompts~\cite{yu2022scaling}. \ours{} outperforms baselines across all metrics.}
        \label{fig:user}
    \end{minipage}%
    \hspace{0.01\textwidth} 
    \begin{minipage}{0.5\textwidth}
        \centering
        \includegraphics[width=\linewidth]{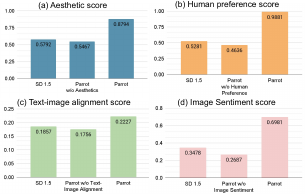}
        \vspace{-2.0em}
        \caption{Ablation study. We perform an ablation study by removing one of quality signals. We observe that each quality signal affects their improvement of (a) aesthetics, (b) human preference score, (c) text-image alignment score, (d) image sentiment score.}
        \label{fig:ablation}
        \vspace{-2.0em}
    \end{minipage}
\end{figure}

\mysubsection{User Study:} We conduct a user study using MTurk~\cite{amt} with generated images from 100 random prompts in the PartiPrompts~\cite{yu2022scaling}. Five models are compared: Stable Diffusion v1.5, DPOK~\cite{fan2023dpok} with an equal weighted sum, Promptist~\cite{hao2022optimizing}, \ours{} without prompt expansion, and \ours{}. Each rater is presented with the original prompt (before expansion) and a set of five generated images, with the image order being randomized. Raters are then tasked with selecting the best image from the group, guided by questions outlined in Table~\ref{tab:mul2}. Each question pertains to a specific quality aspect: aesthetics, human preference, text-image alignment, and image sentiment. For each prompt, 20 rounds of random sampling are conducted and sent to different raters.  The user study results, illustrated in Fig.~\ref{fig:user}, show that \ours{} outperforms other baselines across all dimensions.

\mysubsection{Proportion of non-dominated points:} Using batch size of 256, we observe that in a batch around $20\%$ to $30\%$ are non-dominated points and the proportion of non-dominated points in single batch slightly increase as training proceeds.

\begin{figure*}[t!]
    \centering
    \vspace{0.8em}
    \includegraphics[width=\textwidth]{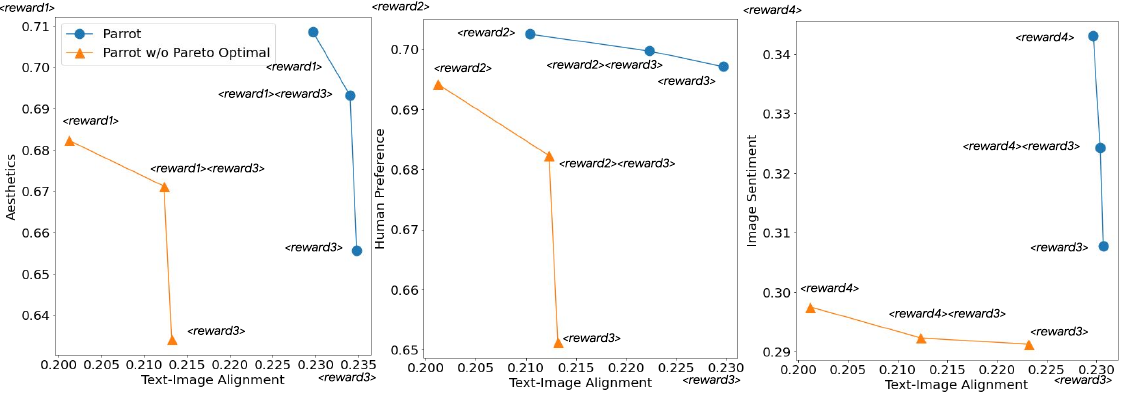}
    \vspace{-2.0em}
    \caption{Comparing Parrot with/without Pareto Optimal. By prepending different reward-specific preferences, \ours{} can change the tradeoff between rewards. Specifically, \textit{``<reward 1>''}, \textit{``<reward 2>''}, \textit{``<reward 3>''}, \textit{``<reward 4>''} indicates aesthetics, human preference, text-image alignment, and image sentiment respectively. With Pareto-optimal selection, each reward increases based on reward-specific preference. }
    \label{fig:d8}
\end{figure*}

\subsection{Ablations}

 \begin{figure}[htp!]
    \centering
    \includegraphics[width=\textwidth]{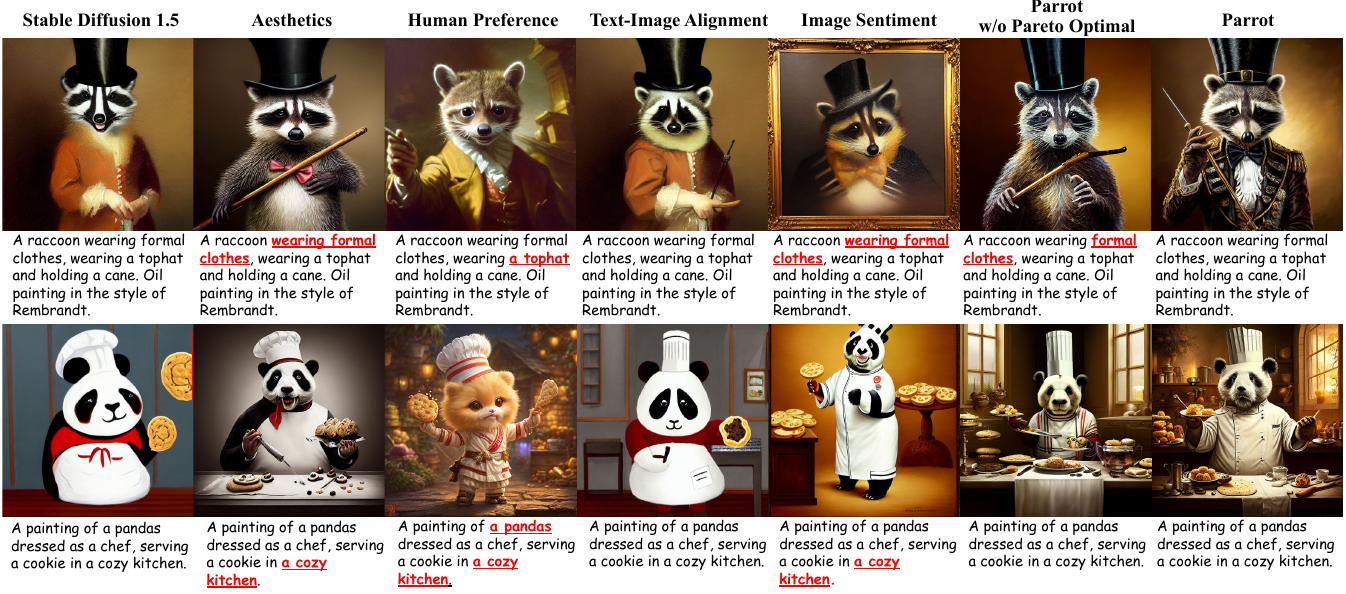}
    \vspace{-2.0em}
    \caption{\textbf{The comparisons of the diffusion fine-tuning between Pareto-optimal multi-reward RL and single reward RL.} We show results with same seed from various methods: Stable Diffusion 1.5~\cite{rombach2022high}~(1st column), T2I model fine-tuned with the aesthetic model~(2nd column), the human preference model~(3rd column), text-image alignment~(4th column), image sentiment~(5th column), Parrot without Pareto-optimal selection~(6th column) and Parrot~(7th column). Parrot is effective to generate acceptable images without sacrificing one of quality signals. For example, T2I model fine-tuned with a single quality signal such as aesthetics, human preference and image sentiment results in text-image misalignment, while our method achieves a balanced visual outcome across multiple criteria. }
    \label{fig:graph}
    
\end{figure}

\mysubsection{Effect of Pareto-optimal Multi-reward RL:} To show the efficacy of Pareto-optimal Multi-reward RL, we conduct an ablation study by removing one reward model at a time. Fig.~\ref{fig:ablation} shows quantitative results using one hundred random text prompts from the Promptist~\cite{hao2022optimizing}. We observe that our training scheme improves multiple target objectives. 

To verify whether \ours{} achieved better trade-off for different reward scores, we generate 1000 images from common animal dataset~\cite{black2023training}, where each text prompt consists of the name of a common animal. As shown in Fig.~\ref{fig:d8}, using only text-image alignment with reward-specific preference~\textit{``$<$reward 3$>$''} generates images with higher text-image alignment score, while using only aesthetic model with reward-specific preference~\textit{``$<$reward 1$>$''} yields images with higher aesthetic score. In the case of using two reward-specific preferences~\textit{``$<$reward 1$>$, $<$reward 3$>$''}, we observe that scores are balanced and show that better results across multiple rewards than~\ours{} without Pareto optimal selection.

Fig~\ref{fig:graph} shows the visual comparison between \ours{}, \ours{} with a single reward, and \ours{} without selecting the batch-wise Pareto-optimal solution. Using a single reward model tends to result in degradation of other rewards, especially text-image alignment. In the third column, results of the first row miss the text~\textit{a tophat} in input prompt, even though the Stable Diffusion result includes that attribute. On the other hand, \ours{} results capture all prompts, improving other quality signals, such as aesthetics, image sentiment and human preference.

\begin{figure}[htp!]
    \centering
    \includegraphics[width=\textwidth]{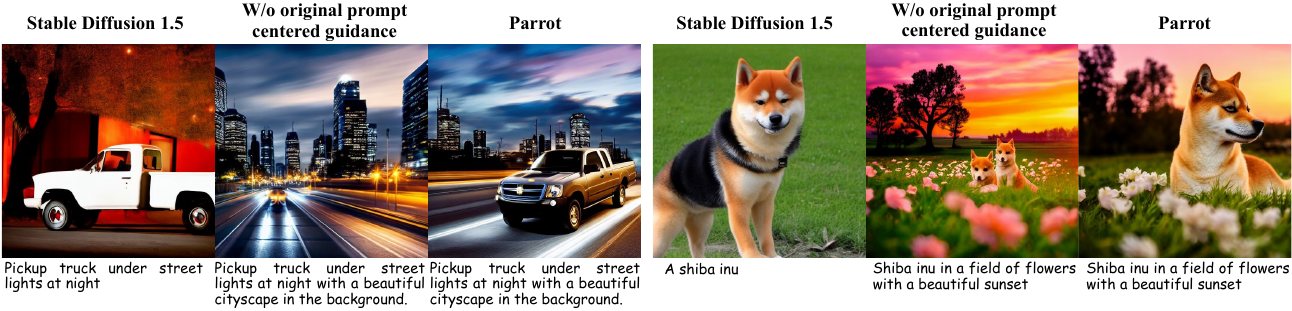}
    \vspace{-2.0em}
    \caption{\textbf{Results of original prompt centered guidance.} As we expand the prompt, the content in the generated image often fades away. This guidance is helpful for keeping the main content of the original prompt.}
    \label{fig:long}
    
\end{figure}

\mysubsection{Effect of Original Prompt Centered Guidance:} Fig~\ref{fig:long} shows the effect of the proposed original prompt-centered guidance.  As evident from the figure, using only the expanded prompt as input often results in the main content being overwhelmed by the added context. For instance, given the original prompt \textit{``A shiba inu"}, the result from the expanded prompt shows a zoomed-out image and the intended main subject (shiba inu) becomes small. The proposed original prompt-centered guidance effectively addresses this issue, generating an image that faithfully captures the input prompt while incorporating visually more pleasing details.


\section{Conclusion and Limitation}
\label{sec:conclusion}

\noindent We propose \ours{}, a novel multi-reward optimization algorithm aimed to improve text-to-image generation by effectively optimizing multiple quality rewards using RL. With batch-wise Pareto-optimal selection, \ours{} adaptively balance the optimization of multiple quality rewards. By applying  \ours{} to jointly finetune both the T2I model and the prompt expansion model, we achieve the generation of higher-quality images with richer details. Additionally, our original prompt centered guidance technique  ensures that the generated image maintains fidelity to the user prompt after prompt expansion during inference. Results from the user study indicate that \ours{} significantly improves the quality of generated images across multiple criteria, including text-image alignment, human preference, aesthetics, and image sentiment. While \ours{} has shown effectiveness in enhancing generated image quality, its efficacy is limited by the quality metrics it relies on. Therefore, advancements of the generated image quality metrics will directly enhance the capabilities of \ours{}. Additionally, \ours{} is adaptable to a broader range of rewards that quantify generated image quality.

\mysubsection{Societal Impact:} Parrot could potentially raise ethical concerns related to the generation of immoral content. This concern stems from the user's ability to influence T2I generation, allowing for the creation of visual content that may be deemed inappropriate. The risk may be tied to the potential biases in reward models inherited from various datasets.

\bibliographystyle{splncs04}
\bibliography{main}

\clearpage
\title{\ours{}: Pareto-optimal Multi-Reward \\  Reinforcement Learning Framework \\ for Text-to-Image Generation \\ Supplementary Material}


\begin{center}
\large
\textbf{\ours{}: Pareto-optimal Multi-Reward \\ Reinforcement Learning Framework for \\ Text-to-Image Generation}\\
Supplementary Material \\
\end{center}

\noindent This supplementary material provides:

\begin{itemize}[leftmargin=*]
\item Sec.~\hyperref[sec:a]{A}: implementation details, including the training details, and details of quantitative experiments. \item Sec.~\hyperref[sec:c]{B}: more ablation studies on original prompt guidance and training scheme of \ours{}.
\item Sec.~\hyperref[sec:d]{C}: more visual examples to show the advancements of \ours{}.

\end{itemize}

\section*{A. Implementation Details}
\label{sec:a}

\mysubsection{Training Details.} We conduct our experiments with Jax implementation of Stable Diffusion 1.5~\cite{rombach2022high}. In terms of diffusion-based RL, we sample 256 images per RL-tuning iteration. For policy gradient updates, we accumulate gradients across all denoising timesteps. Our experiments employ a small range of gradient clip $10^{-4}$. We keep negative prompt as null text.

\mysubsection{Details of Quantitative Experiments.} From Parti~\cite{yu2022scaling} prompts, we generate images of dimensions $512 \times 512$. In all experiments in the main paper, we apply reward-specific preference expressed as~\textit{``$<$reward 1$>$, $<$reward 2$>$, $<$reward 3$>$, $<$reward 4$>$''}, which is optional to select one or several rewards. Reward models are aesthetics, human preference, text-image alignment and image sentiment. During the inference stage, the guidance scale is also set as 5.0 for the diffusion sampling process. We employ the AdamW~\cite{kingma2014adam} as optimizer with $\beta_1=0.9$, $\beta_2=0.999$ and a weight decay of 0.1. 

\section*{B. Prompt Guidance and Training Scheme}

\label{sec:c}

\begin{figure*}[t!]
    \centering
    \includegraphics[width=\textwidth]{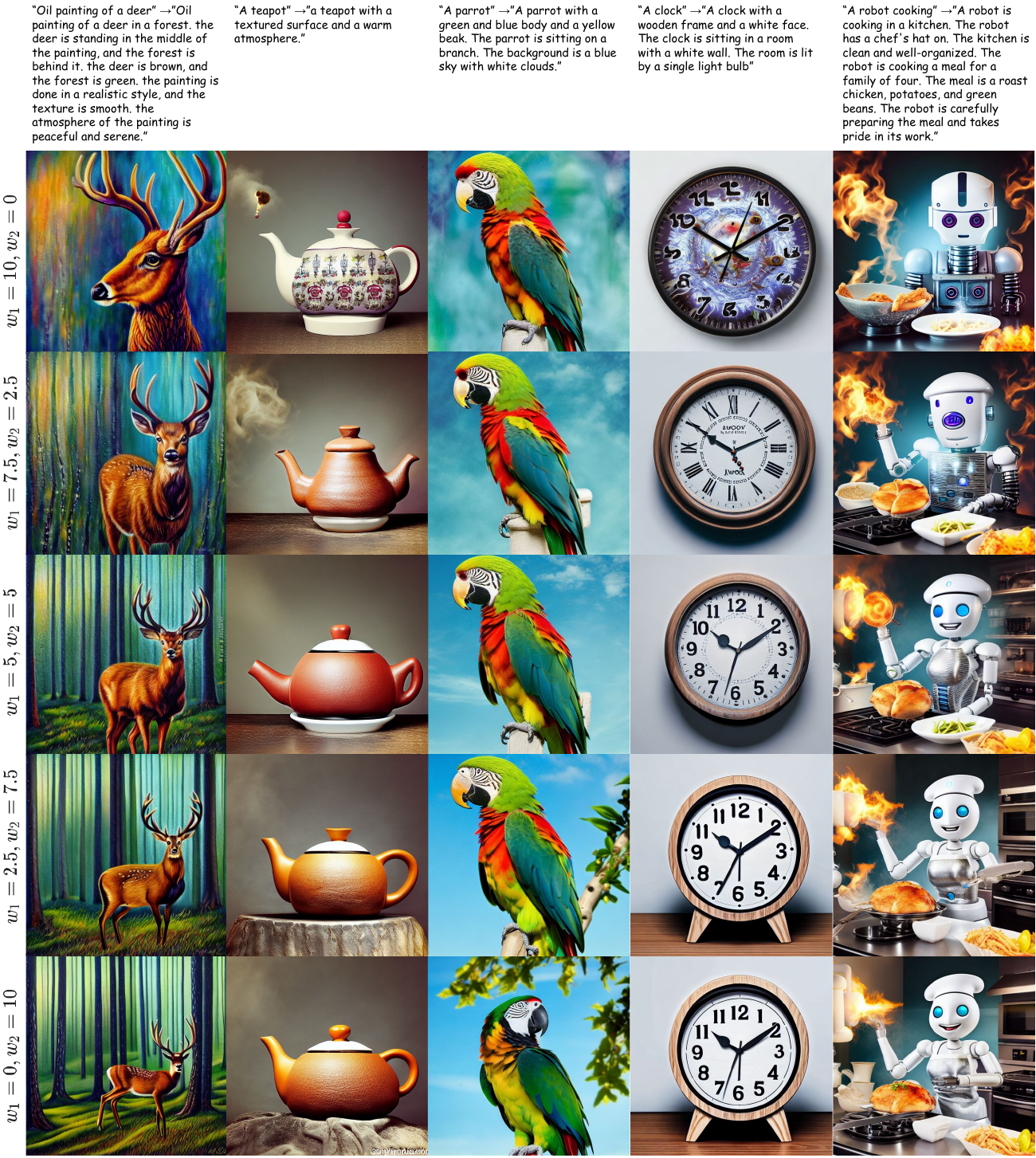}
    \caption{Original prompt centered guidance. We present visual comparison of 5 different pairs of $w_1$ and $w_2$ to demonstrate the effectiveness of guidance scales. For all experiments, we assign $w_1=5$ and $w_2=5$ (3rd row) for the best performance. }
    \label{fig:d10}
\end{figure*}

\mysubsection{Original Prompt Guidance.} In Fig.~\ref{fig:d10}, we provide additional ablation study of original prompt centered guidance by adjusting the guidance scale $w_1$ and $w_2$, which determines visual changes from original prompt and expanded prompt, respectively. We assigned 5 to both $w1$ and $w2$ in most of our experiments, which shows better text-image alignment performance in Fig.~\ref{fig:d10}.

\begin{figure*}[t!]
    \centering
    \includegraphics[width=\textwidth]{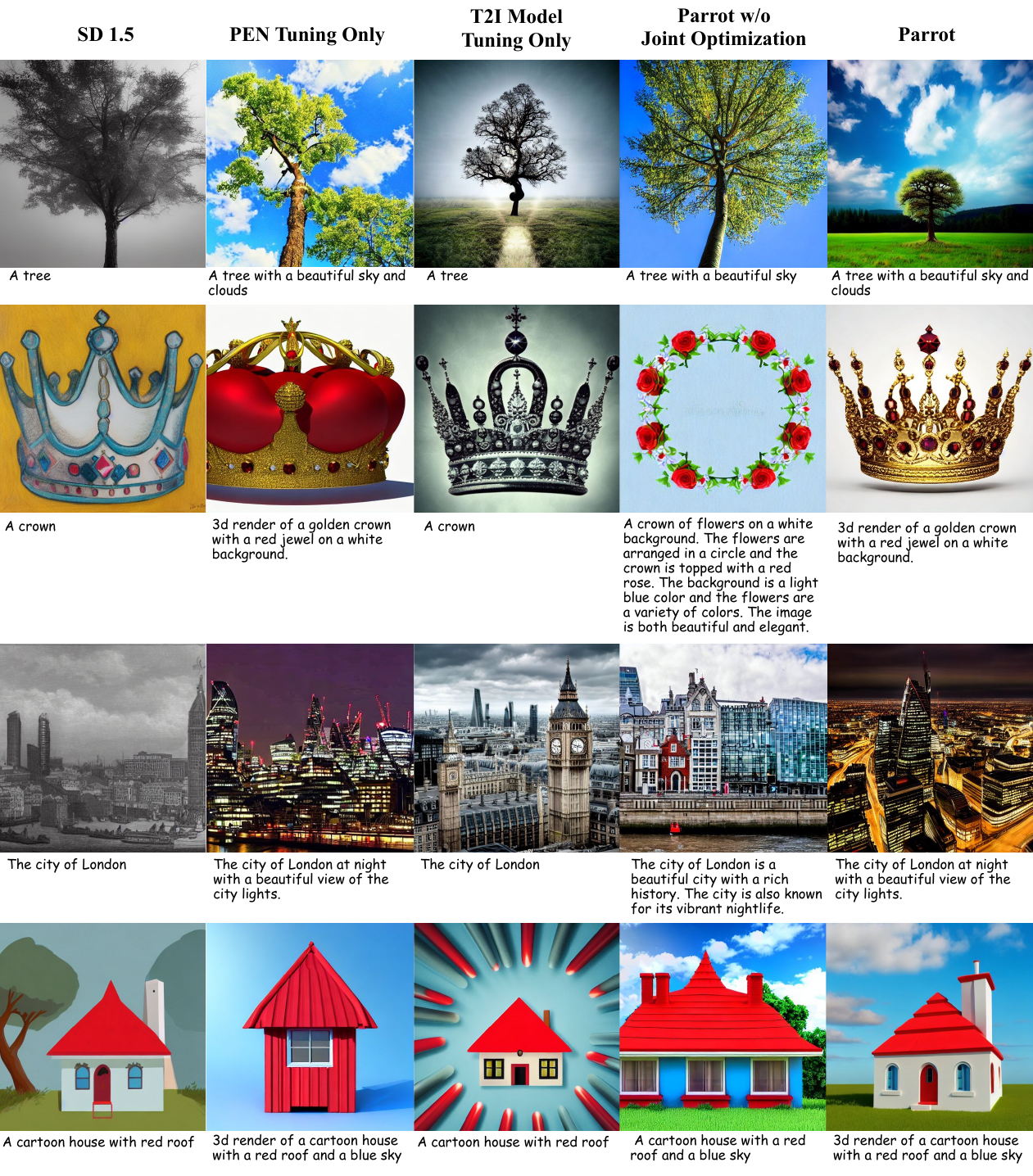}
    \caption{Visual examples of \textbf{\ours{}} under different settings. From left to right, we provide results of Stable Diffusion 1.5~\cite{rombach2022high}, the only fine-tuned PEN, the only fine-tuned T2I diffusion model, the \textbf{\ours{}} without joint optimization, and the \textbf{\ours{}}. }
    \label{fig:d5}
\end{figure*}

\mysubsection{Ablation Study on Training Scheme.} Fig.~\ref{fig:d5} provides ablation study comparing variation of the \ours{}: Stable Diffusion, the prompt expansion network~(PEN) tuning only, T2I model fine-tuning only, \ours{} without joint optimization, and \ours{}. In the second column, without fine-tuning the T2I diffusion model does not lead to significant improvements in terms of texture and composition. Furthermore, the third column demonstrates that fine-tuning the diffusion model enhances texture and perspective, yet this improvement is hindered by the limited information in the text prompt. We also observe that the quality of images from joint optimization surpasses that of combining decoupled generative models.

\section*{C. More Visual Examples}

\label{sec:d}

\begin{figure*}[t!]
    \centering
    \includegraphics[width=\textwidth]{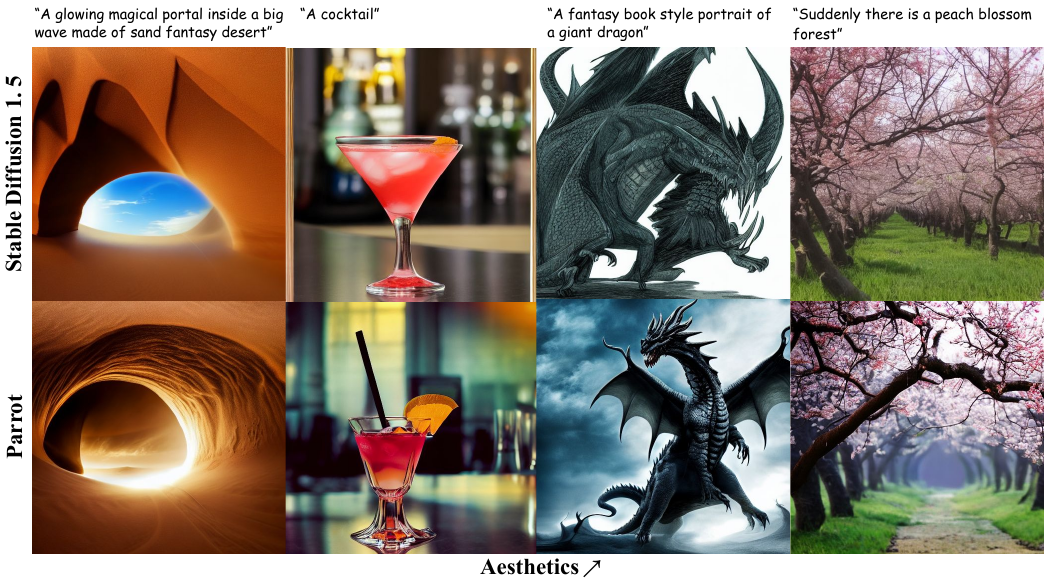}
    \caption{More Examples of aesthetics improvement from the \textbf{\ours{}}. Given the text prompt, we generate images with Stable Diffusion and ~\textbf{\ours{}}. After fine-tuning, the \textbf{\ours{}} alleviates quality issues such as poor composition (e.g. bad cropping), misalignment with the user input (e.g. missing objects), or generally less aesthetic pleasing.}
    \label{fig:d1}
\end{figure*}

\begin{figure*}[t!]
    \centering
    \includegraphics[width=\textwidth]{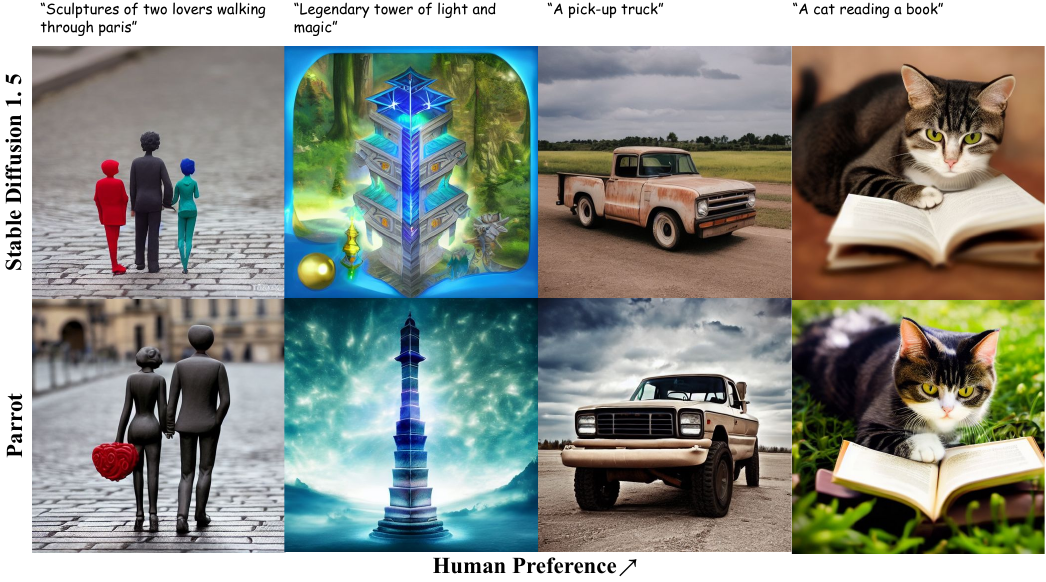}
    \caption{More Examples of human preference improvement from the \textbf{\ours{}}. Given the text prompt, we generate images with Stable Diffusion 1.5~\cite{rombach2022high} and ~\textbf{\ours{}}.}
    \label{fig:d2}
\end{figure*}

\begin{figure*}[t!]
    \centering
    \includegraphics[width=\textwidth]{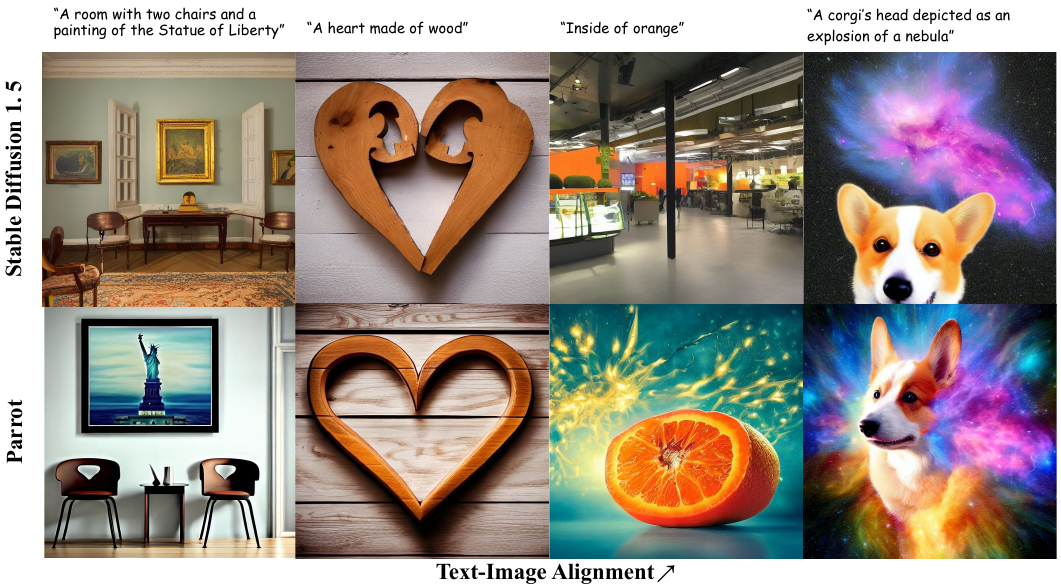}
    \caption{More examples of text-image alignment improvement from the \textbf{\ours{}}. Given the text prompt, we generate images with the Stable Diffusion  1.5~\cite{rombach2022high} and the~\textbf{\ours{}}.}
    \label{fig:d4}
\end{figure*}

\begin{figure*}[t!]
    \centering
    \includegraphics[width=\textwidth]{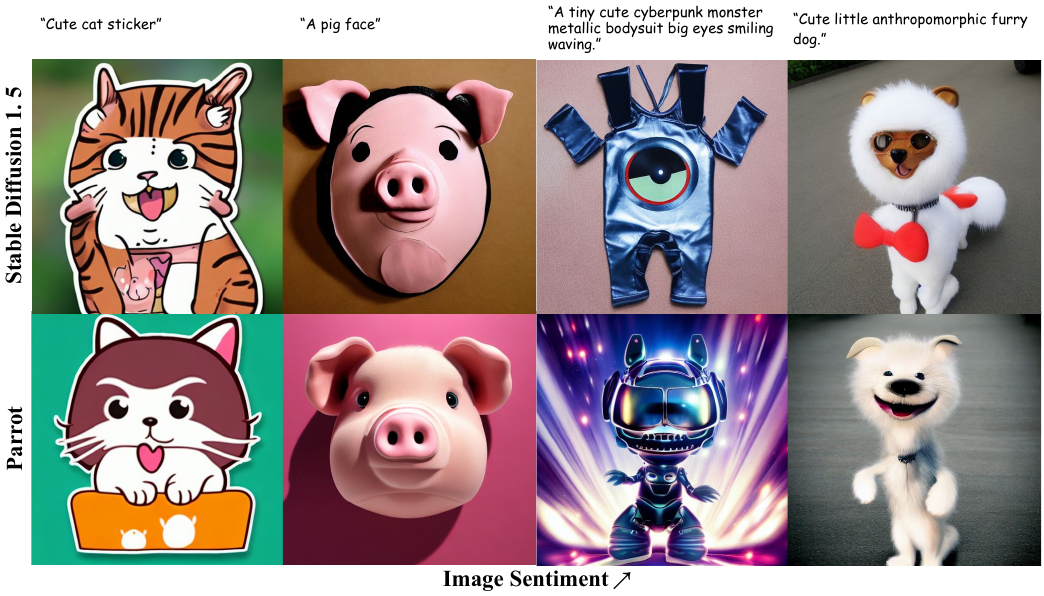}
    \vspace{-2.0em}
    \caption{More examples of image sentiment improvement from the \textbf{\ours{}}. Given the text prompt, we generate images with the Stable Diffusion 1.5~\cite{rombach2022high} and the~\textbf{\ours{}}.}
    \vspace{-1.5em}
    \label{fig:d3}
\end{figure*}

\begin{figure*}[t!]
    \centering
    \includegraphics[width=\textwidth]{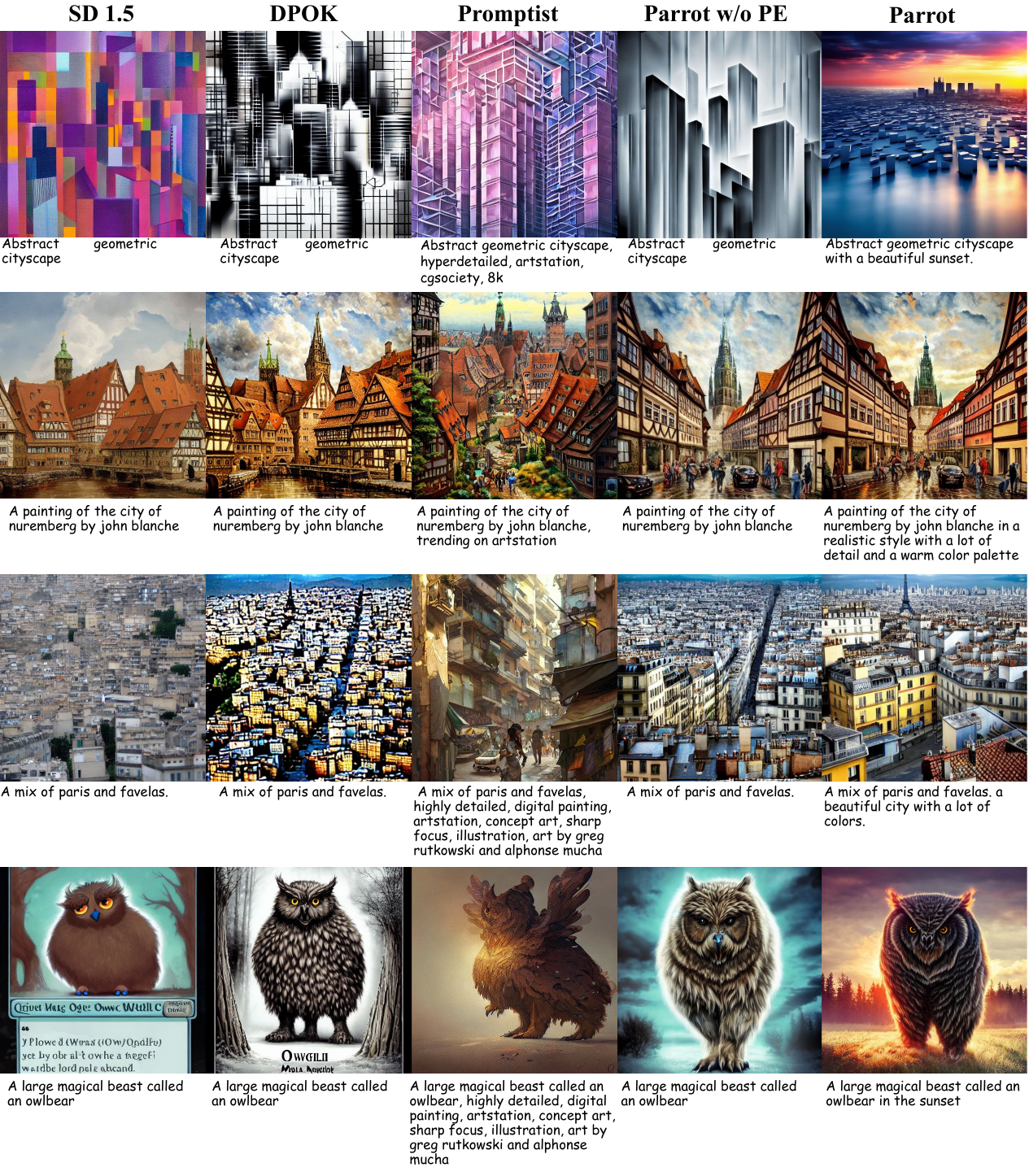}
    \vspace{-2.0em}
    \caption{More results from the \textbf{\ours{}} and baselines: Stable Diffusion 1.5~\cite{rombach2022high}, DPOK~\cite{fan2023dpok} with weighted sum, Promptist~\cite{fan2023optimizing}, \ours{} without prompt expansion, and \ours{}.}
    \vspace{-1.5em}
    \label{fig:d6}
\end{figure*}

\begin{figure*}[t!]
    \centering
    \includegraphics[width=\textwidth]{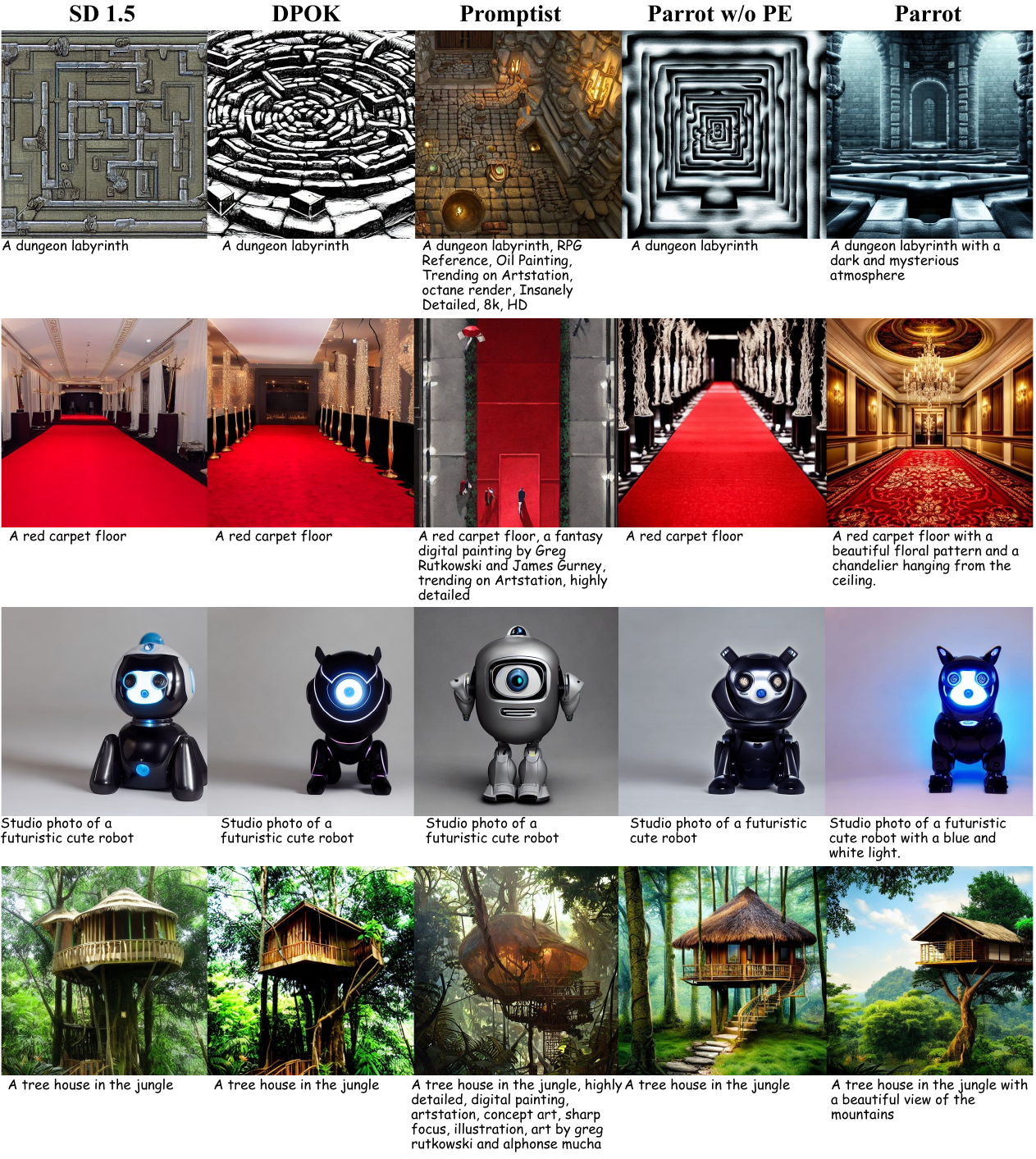}
    \vspace{-2.0em}
    \caption{More results from the \textbf{\ours{}} and baselines: Stable Diffusion 1.5~\cite{rombach2022high}, DPOK~\cite{fan2023dpok} with weighted sum, Promptist~\cite{fan2023optimizing}, \ours{} without prompt expansion, and \ours{}.}
    \vspace{-1.5em}
    \label{fig:d7}
\end{figure*}

We show additional visual examples of \ours{} in~\cref{fig:d1,fig:d2,fig:d3,fig:d4,fig:d6,fig:d7}. Note that generated images from \ours{} are improved across multiple-criteria. Fig.~\ref{fig:d1} highlights examples where the \ours{} brings improvements in aesthetics. For example, \ours{} effectively addresses issues such as poor cropping in the fourth column and improves color in the fifth column. Fig.~\ref{fig:d2} presents examples of images with improved human preference score generated by \ours{}. In Fig.~\ref{fig:d4}, we provide examples of improved text-image alignment achieved by \ours{}. Fig.~\ref{fig:d3} shows examples where \ours{} enhances image sentiment, producing emotionally rich images. 

Finally, additional comparison results between diffusion-based RL baselines are described in Fig.~\ref{fig:d6}, and Fig.~\ref{fig:d7}. Diffusion-based RL baselines are listed: Stable Diffusion 1.5~\cite{rombach2022high}, DPOK~\cite{fan2023dpok} with weighted sum of multiple reward scores, Promptist~\cite{fan2023optimizing}, \ours{} without prompt expansion, and \ours{}. For \ours{} without prompt expansion, we only take original prompt as input.

\end{document}